\documentclass[runningheads]{llncs}
\usepackage{graphicx}
\usepackage{amsmath,amssymb}
\newcommand{\etal}{\textit{et al}}
\begin{document}

\title{ReCoNet: Real-time Coherent Video Style Transfer Network} 
\titlerunning{ReCoNet} 


\author{Chang Gao$^\star$\inst{1} \and
Derun Gu$^\star$\inst{1} \and
Fangjun Zhang\thanks{Joint first authors.}\inst{1} \and
Yizhou Yu\inst{1,2}}
%

\authorrunning{Gao et al.} 


\institute{The University of Hong Kong\\
\email{\{u3514174,greatway,u3514241\}@connect.hku.hk}\\
Deepwise AI Lab\\
\email{yizhouy@acm.org}}


\maketitle

\begin{abstract}
Image style transfer models based on convolutional neural networks usually suffer from high temporal inconsistency when applied to videos. Some video style transfer models have been proposed to improve temporal consistency, yet they fail to guarantee fast processing speed, nice perceptual style quality and high temporal consistency at the same time. In this paper, we propose a novel real-time video style transfer model, ReCoNet, which can generate temporally coherent style transfer videos while maintaining favorable perceptual styles. A novel luminance warping constraint is added to the temporal loss at the output level to capture luminance changes between consecutive frames and increase stylization stability under illumination effects. We also propose a novel feature-map-level temporal loss to further enhance temporal consistency on traceable objects. Experimental results indicate that our model exhibits outstanding performance both qualitatively and quantitatively.

\keywords{Video style transfer \and Optical flow \and Real-time processing.}
\end{abstract}
\section{Introduction}

\begin{figure}[t]
\centering
\includegraphics[width=10cm]{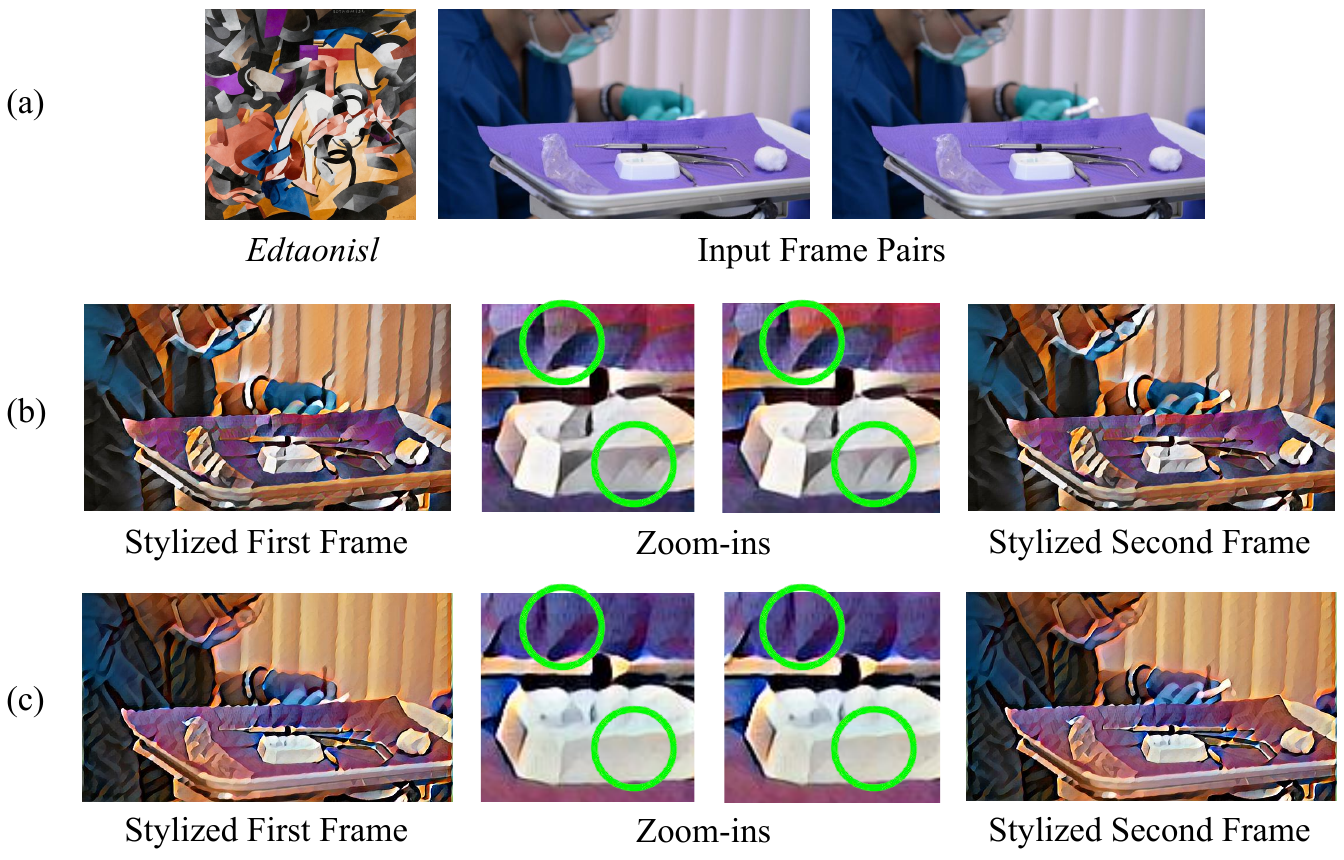}
\caption{Temporal inconsistency in video style transfer. (a) The style target \textit{Edtaonisl} (Francis Picabia, 1913) and two consecutive video frames from Videvo.net~\cite{videvonet} (b) Style transfer results by Chen~\etal~\cite{chen2017coherent} (c) Style transfer results by ReCoNet. The circled regions show that our model can better suppress temporal inconsistency, while Chen~\etal's model generates inconsistent color and noticeable flickering effects}
\label{coherence}
\end{figure}

As a natural extension of image style transfer, video style transfer has recently gained interests among researchers~\cite{chen2017coherent,gupta2017characterizing,huang2017real,ruder2016artistic,ruder2018artistic,anderson2016deepmovie,chen2018stereoscopic}. Although some image style transfer methods~\cite{johnson2016perceptual,dumoulin2016learned} have achieved real-time processing speed, i.e. around or above 24 frames per second (FPS), one of the most critical issues in their stylization results is high temporal inconsistency. Temporal inconsistency, or sometimes called incoherence, can be observed visually as flickering between consecutive stylized frames and inconsistent stylization of moving objects~\cite{chen2017coherent}. Figure \ref{coherence}(a)(b) demonstrate temporal inconsistency in video style transfer.

To mitigate this effect, optimization methods guided by optical flows and occlusion masks were proposed~\cite{anderson2016deepmovie,ruder2016artistic}. Although these methods can generate smooth and coherent stylized videos, it generally takes several minutes to process each video frame due to optimization on the fly. Some recent models~\cite{chen2017coherent,gupta2017characterizing,huang2017real,ruder2018artistic} improved the speed of video style transfer using optical flows and occlusion masks explicitly or implicitly, yet they failed to guarantee real-time processing speed, nice perceptual style quality, and coherent stylization at the same time.

In this paper, we propose ReCoNet, a real-time coherent video style transfer network as a solution to the aforementioned problem. ReCoNet is a feed-forward neural network which can generate coherent stylized videos with rich artistic strokes and textures in real-time speed. It stylizes videos frame by frame through an encoder and a decoder, and uses a VGG loss network~\cite{johnson2016perceptual,simonyan2014very} to capture the perceptual style of the transfer target. It also incorporates optical flows and occlusion masks as guidance in its temporal loss to maintain temporal consistency between consecutive frames, and the effects can be observed in Figure \ref{coherence}(c). In the inference stage, ReCoNet can run far above the real-time standard on modern GPUs due to its lightweight and feed-forward network design.

We find that the brightness constancy assumption~\cite{horn1974determining} in optical flow estimation may not strictly hold in real-world videos and animations, and there exist luminance differences on traceable pixels between consecutive image frames. Such luminance differences cannot be captured by temporal losses purely based on optical flows. To consider the luminance difference, we further introduce a luminance warping constraint in our temporal loss.

From stylization results of previous methods~\cite{chen2017coherent,gupta2017characterizing,huang2017real,ruder2016artistic,ruder2018artistic}, we have also observed instability such as different color appearances of the same moving object in consecutive frames. With the intuition that the same object should possess the same features in high-level feature maps, we apply a feature-map-level temporal loss to our encoder. This further improves temporal consistency of our model.

In summary, there exist the following contributions in our paper:
\begin{itemize}
\item Our model highly incorporates perceptual style and temporal consistency in the stylized video. With a new feed-forward network design, it can achieve an inference speed over 200 FPS on a single modern GPU. Our model can reproduce various artistic styles on videos with stable results.
\item We first propose a luminance warping constraint in the output-level temporal loss to specifically consider luminance changes of traceable pixels in the input video. This constraint can improve stylizing stability in areas with illumination effects and help suppress overall temporal inconsistency.
\item We first propose a feature-map-level temporal loss to penalize variations in high-level features of the same object in consecutive frames. This improves stylizing stability of traceable objects in video scenes.
\end{itemize}

In this paper, related work for image and video style transfer will be first reviewed in Section 2. Detailed motivations, network architecture, and loss functions will be presented in Section 3. In Section 4, the experiment results will be reported and analyzed, where our model shows outstanding performance.

\section{Related Work} \label{sec:related_work}

Gatys~\etal~\cite{gatys2015neural,gatys2016image} first developed a neural algorithm for automatic image style transfer, which refines a random noise to a stylized image iteratively constrained by a content loss and a style loss. This method inspired many later image style transfer models~\cite{johnson2016perceptual,dumoulin2016learned,selim2016painting,champandard2016semantic,ulyanov2016texture,luan2017deep,gatys2017controlling,chen2017stylebank,liao2017visual}. One of the most successful successor is the feed-forward perceptual losses model proposed by Johnson~\etal~\cite{johnson2016perceptual}, using a pre-trained VGG network~\cite{simonyan2014very} to compute perceptual losses. Although their model has achieved both preferable perceptual quality and near real-time inference speed, severe flickering artifacts can be observed when applying this method frame by frame to videos since temporal stability is not considered. Afterwards, Anderson~\etal~\cite{anderson2016deepmovie} and Ruder~\etal~\cite{ruder2016artistic} introduced a temporal loss function in video stylization as an explicit consideration of temporal consistency. The temporal loss is involved with optical flows and occlusion masks and is iteratively optimized for each frame until the loss converges. However, it generally takes several minutes for their models to process each video frame, which is not applicable for real-time usage. Although Ruder~\etal~\cite{ruder2018artistic} later accelerated the inference speed, their stylization still runs far below the real-time standard.

To obtain a consistent and fast video style transfer method, some real-time or near real-time models have recently been developed. Chen~\etal~\cite{chen2017coherent,chen2018stereoscopic} proposed a recurrent model that uses feature maps of the previous frame in addition to the input consecutive frames, and involves explicit optical flows warping on feature maps in both training and inference stages. Since this model requires optical flow estimation by \textit{FlowNetS}~\cite{dosovitskiy2015flownet} in the inference stage, its inference speed barely reaches real-time level and the temporal consistency is susceptible to errors in optical flow estimation. Gupta~\etal~\cite{gupta2017characterizing} also proposed a 
recurrent model which takes an additional stylized previous frame as the input. Although their model performs similarly to Chen~\etal's model in terms of temporal consistency, it suffers from transparency issues and still barely reaches real-time inference speed. Using a feed-forward network design, Huang~\etal~\cite{huang2017real} proposed a model similar to the perceptual losses model~\cite{johnson2016perceptual} with an additional temporal loss. This model is faster since it neither estimates optical flows nor uses information of previous frames in the inference stage. However, Huang~\etal's model calculates the content loss from a deeper layer $relu4\_2$, which is hard to capture low-level features. Strokes and textures are also weakened in their stylization results due to a low weight ratio between perceptual losses and the temporal loss.

Noticing strengths and weaknesses of these models, we propose several improvements in ReCoNet. Compared with Chen~\etal~\cite{chen2017coherent}'s model, our model does not estimate optical flows but involves ground-truth optical flows only in loss calculation in the training stage. This can avoid optical flow prediction errors and accelerate inference speed. Meanwhile, our model can render style patterns and textures much more conspicuously than Huang~\etal~\cite{huang2017real}'s model, which could only generate minor visual patterns and strokes besides color adjustment. Our lightweight and feed-forward network can run faster than all video stylization models mentioned above~\cite{chen2017coherent,gupta2017characterizing,huang2017real,ruder2016artistic,ruder2018artistic,anderson2016deepmovie}.

\section{Method}

\begin{figure}[t]
\centering
\includegraphics[width=12cm]{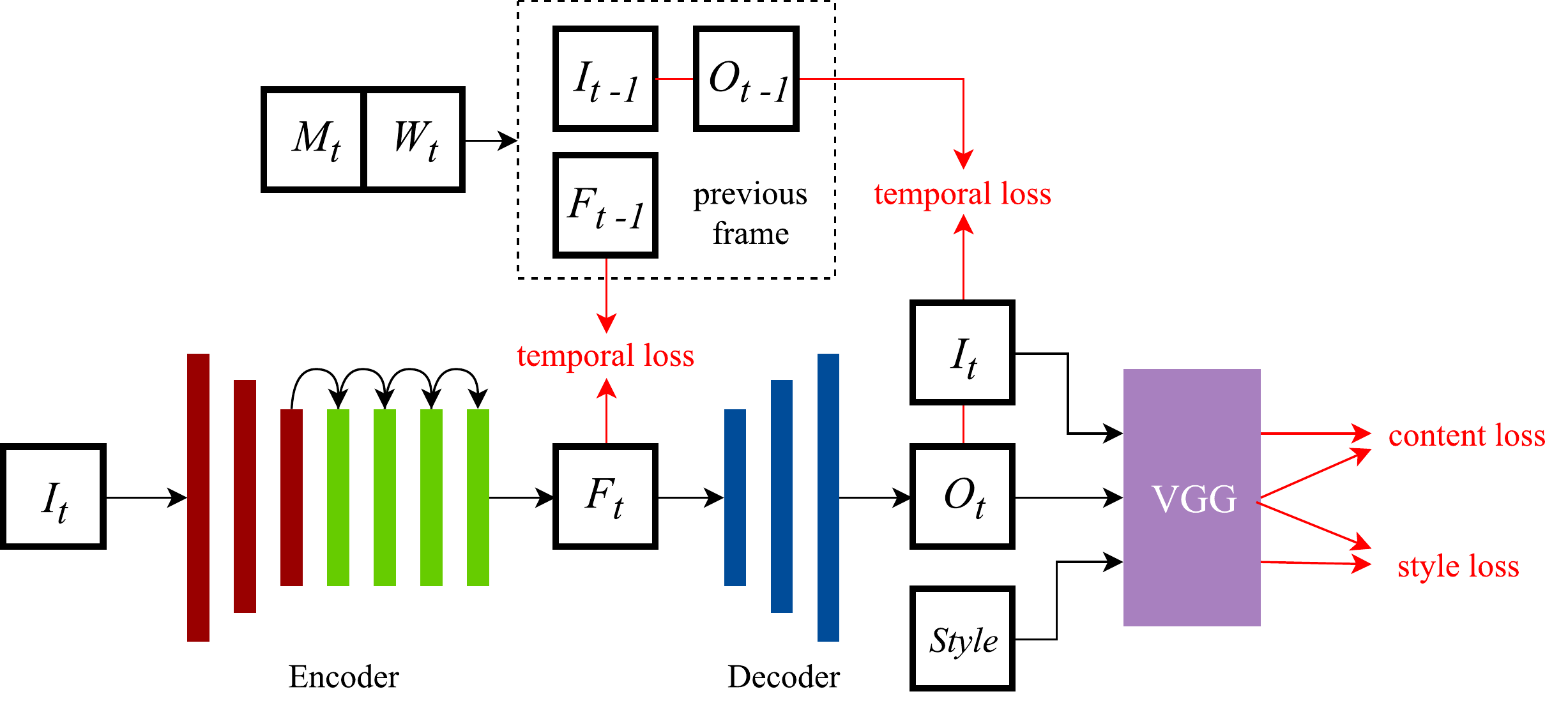}
\caption{The pipeline of ReCoNet. $I_t, F_t, O_t$ denote the input image, encoded feature maps, and stylized output image at time frame $t$. $M_t$ and $W_t$ denote the occlusion mask and the optical flow between time frames $t-1$ and $t$. $Style$ denotes the artistic style image. The dashed box represents the prediction results of the previous frame, which will only be used in the training process. Red arrows and texts denote loss functions}
\label{pipeline}
\end{figure}

The training pipeline of ReCoNet is shown in Figure \ref{pipeline}. ReCoNet consists of three modules: an encoder that converts input image frames to encoded feature maps, a decoder that generates stylized images from feature maps, and a VGG-16~\cite{simonyan2014very} loss network to compute the perceptual losses. Additionally, a multi-level temporal loss is added to the output of encoder and the output of decoder to reduce temporal incoherence. In the inference stage, only the encoder and the decoder will be used to stylize videos frame by frame.

\subsection{Motivation} \label{sec:motivation}

\subsubsection{Luminance Difference}

In real-world videos, the luminance and color appearances can be different on the same object in consecutive frames due to illumination effects. In such cases, the data does not satisfy the assumption known as brightness constancy constraint~\cite{horn1974determining}, and direct optical flow warping will ignore luminance changes in traceable pixels~\cite{dosovitskiy2015flownet,weinzaepfel2013deepflow,ilg2017flownet}. In animations, many datasets use the albedo pass to calculate ground-truth optical flows but later add illuminations including smooth shading and specular reflections to the final image frames, such as MPI Sintel Dataset~\cite{butler2012naturalistic}. This also results in differences on luminance and color appearances.

\begin{figure}[t]
\centering
\includegraphics[width=12cm]{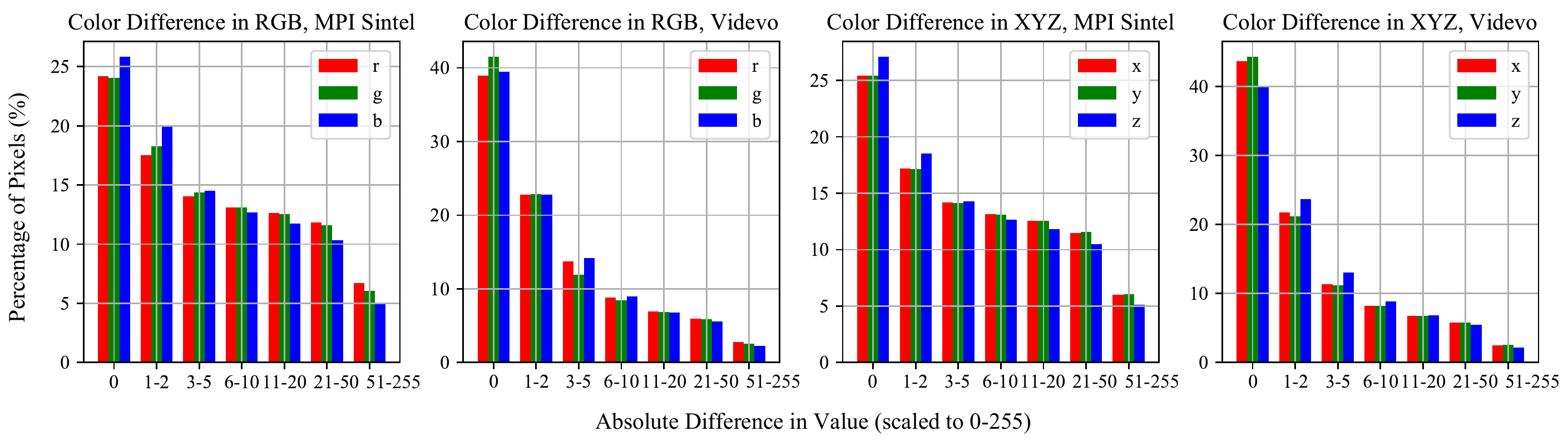}
\caption{Histograms of temporal warping error in different datasets and color spaces}
\label{colorspace}
\end{figure}

To further examine the illumination difference, we computed the absolute value of temporal warping error $I_t - W_t(I_{t-1})$ over MPI Sintel Dataset and 50 real-world videos download from Videvo.net~\cite{videvonet}, where $W$ is the forward optical flow and $I$ is the input image frame. We used \textit{FlowNet2}~\cite{ilg2017flownet} to calculate optical flows and the method of Sundaram~\etal~\cite{sundaram2010dense} to obtain occlusion masks for downloaded videos. Figure \ref{colorspace} demonstrates the histograms of temporal warping error in both RGB and XYZ color space. We can draw two conclusions based on the results. First, RGB channels share similar warping error distributions. There is no bias of changes in color channel. Second, despite changes in relative luminance channel Y, the chromaticity channels X and Z in XYZ color space also contribute to the total inter-frame difference. However, since there is no exact guideline of chromaticity mapping in a particular style, we mainly consider luminance difference in our temporal loss.

Based on our findings, we propose a novel luminance constraint in our temporal loss to encourage the stylized frames to have the same luminance changes as the input frames. This can reduce unstable color changes under illumination effects and improve temporal consistency of the stylized frames. Experiments in Section \ref{sec:ablation} show that this new constraint can bring significant improvements to the output perceptual quality and temporal stability.

\subsubsection{Feature-map-level Temporal Loss}

Another new loss function we propose for feature-map-level consistency is based on the intuition that the same object should preserve the same representation in high-level feature maps. Although warping frames directly at the output level may not be accurate due to illuminations, the same method can be very suitable at the feature-map level as examined by Chen~\etal~\cite{chen2017coherent}. We use ground-truth optical flows and occlusion masks to calculate feature-map-level temporal loss between the warped feature maps and the current ones. Experiments in Section \ref{sec:ablation} show that this new loss can improve stylization consistency on the same object in consecutive frames. 

\subsection{Network Architecture}

ReCoNet adopts a pure CNN-based design. Compared to feed-forward networks in literature~\cite{huang2017real,johnson2016perceptual}, we separate the whole network to an encoder and a decoder for different purposes. The encoder is designed to encode image frames to feature maps with aggregated perceptual information, and the feature-map-level temporal loss is computed on its output. The decoder is designed to decode feature maps to a stylized image where we compute the output-level temporal loss. Table \ref{encoder_decoder} shows our encoder and decoder design. There are three convolutional layers and four residual blocks~\cite{he2016deep} in the encoder, and two up-sampling convolutional layers with a final convolutional layer in the decoder. We use an up-sample layer and a convolutional layer instead of one traditional deconvolutional layer in the decoder to reduce checkerboard artifacts~\cite{odena2016deconvolution}. We adopt instance normalization~\cite{ulyanov2016instance} after each convolution process to attain better stylization quality. Reflection padding is used at each convolutional layer.

The loss network is a VGG-16 network~\cite{simonyan2014very} pre-trained on the ImageNet dataset~\cite{deng2009imagenet}. For each iteration, the VGG-16 network processes each of the input image frame, output image frame and style target independently. The content and style losses are then computed based on the generated image features.

\setlength{\tabcolsep}{4pt}
\begin{table}[t]
\caption{Network layer specification. Layer and output sizes are denoted as channel $\times$ height $\times$ width. \textit{Conv, Res, InsNorm, ReLU, Tanh} denote convolutional layer, residual block~\cite{he2016deep}, instance normalization layer~\cite{ulyanov2016instance}, ReLU activation layer~\cite{nair2010rectified}, and Tanh activation layer respectively}

\begin{center}
\begin{tabular}{| c | c| c | c |}
\hline Layer & Layer Size & Stride & Output Size\\\hline 
\multicolumn{4}{|c|}{Encoder}\\\hline 
Input & & & $3 \times 640 \times 360$\\
Conv + InsNorm + ReLU & $48 \times 9 \times 9$ & 1 & $48 \times 640 \times 360$\\
Conv + InsNorm + ReLU & $96 \times 3 \times 3$ & 2 & $96 \times 320 \times 180$\\
Conv + InsNorm + ReLU & $192 \times 3 \times 3$ & 2 & $192 \times 160 \times 90$\\
(Res + InsNorm + ReLU ) $\times 4$ & $192 \times 3 \times 3$ & 1 & $192 \times 160 \times 90$\\\hline 
\multicolumn{4}{|c|}{Decoder}\\\hline 
Up-sample & & 1/2 & $192 \times 320 \times 180$\\
Conv + InsNorm + ReLU & $96 \times 3 \times 3$ & 1 & $96 \times 320 \times 180$\\
Up-sample & & 1/2 & $96 \times 640 \times 360$\\
Conv + InsNorm + ReLU & $48 \times 3 \times 3$ & 1 & $48 \times 640 \times 360$\\
Conv + Tanh & $3 \times 9 \times 9$ & 1 & $3 \times 640 \times 360$\\\hline 
\end{tabular}
\end{center}
\label{encoder_decoder}
\end{table}
\setlength{\tabcolsep}{1.4pt}

\subsection{Loss functions}

Our multi-level temporal loss design focuses on temporal coherence at both high-level feature maps and the final stylized output. At the feature-map level, a strict optical flow warping is adopted to achieve temporal consistency of traceable pixels in high-level features. At the output level, an optical flow warping with a luminance constraint is used to simulate both the movements and luminance changes of traceable pixels. The perceptual losses design is inherited from the perceptual losses model~\cite{johnson2016perceptual}.

A two-frame synergic training mechanism~\cite{huang2017real} is used in the training stage. For each iteration, the network generates feature maps and stylized output of the first image frame and the second image frame in two runs. Then, the temporal losses are computed using the feature maps and stylized output of both frames, and the perceptual losses are computed on each frame independently and summed up. Note again that in the inference stage, only one image frame will be processed by the network in a single run.

\subsubsection{Output-level Temporal loss}
The temporal losses in previous works~\cite{chen2017coherent,gupta2017characterizing,huang2017real,ruder2016artistic,ruder2018artistic} usually ignore changes in luminance of traceable pixels. Taking this issue into account, the \textit{relative luminance} $Y=0.2126R+0.7152G+0.0722B$, same as Y in XYZ color space, is added as a warping constraint for all channels in RGB color space:
\begin{equation} \label{eq:temp_o}
\mathcal{L}_{temp, o}(t-1, t) = \sum_c \frac{1}{D} M_t \lVert (O_t - W_t(O_{t-1}))_c - (I_t - W_t(I_{t-1}))_Y \rVert^2
\end{equation}
where $c \in [R, G, B]$ is each of the RGB channels of the image, $Y$ the relative luminance channel, $O_{t-1}$ and $O_t$ the stylized images for previous and current input frames respectively, $I_{t-1}$ and $I_t$ the previous and current input frames respectively, $W_t$ the ground-truth forward optical flow, $M_t$ the ground-truth forward occlusion mask (1 at traceable pixels or 0 at untraceable pixels), $D = H \times W$ the multiplication of height $H$ and width $W$ of the input/output image. We apply the relative luminance warping constraint to each RGB channel equally based on the ``no bias'' conclusion in Section \ref{sec:motivation}. Section \ref{sec:ablation} further discusses different choices of the luminance constraint and the output-level temporal loss.

\subsubsection{Feature-map-level Temporal loss}
The feature-map-level temporal loss penalizes temporal inconsistency on the encoded feature maps between two consecutive input image frames:
\begin{equation}
\mathcal{L}_{temp, f}(t-1, t) = \frac{1}{D} M_t \lVert F_t - W_t(F_{t-1}) \rVert^2
\end{equation}
where $F_{t-1}$ and $F_t$ are the feature maps outputted by the encoder for previous and current input frames respectively, $W_t$ and $M_t$ the ground-truth forward optical flow and occlusion mask downscaled to the size of feature maps, $D = C \times H \times W$ the multiplication of channel size $C$, image height $H$ and image width $W$ of the encoded feature maps $F$. We use downscaled optical flows and occlusion masks to simulate temporal motions in high-level features.

\subsubsection{Perceptual Losses}
We adopt the content loss $\mathcal{L}_{content}(t)$, the style loss $\mathcal{L}_{style}(t)$ and the total variation regularizer $\mathcal{L}_{tv}(t)$ in the perceptual losses model~\cite{johnson2016perceptual} for each time frame $t$. The content loss and the style loss utilize feature maps at $relu3\_3$ layer and $[relu1\_2, relu2\_2, relu3\_3, relu4\_3]$ layers respectively.

\subsubsection{Summary}
The final loss function for the two-frame synergic training is:
\begin{multline}
\mathcal{L}(t-1, t)= \sum_{i \in \{t-1, t\}}(\alpha \mathcal{L}_{content}(i) + \beta \mathcal{L}_{style}(i) + \gamma \mathcal{L}_{tv}(i))\\ + \lambda_f \mathcal{L}_{temp, f}(t-1, t) + \lambda_o \mathcal{L}_{temp, o}(t-1, t)
\end{multline}
where $\alpha, \beta, \gamma, \lambda_f$ and $\lambda_o$ are hyper-parameters for the training process.

\section{Experiments}

\subsection{Implementation Details}

\begin{figure}
\centering
\includegraphics[width=1.0\textwidth]{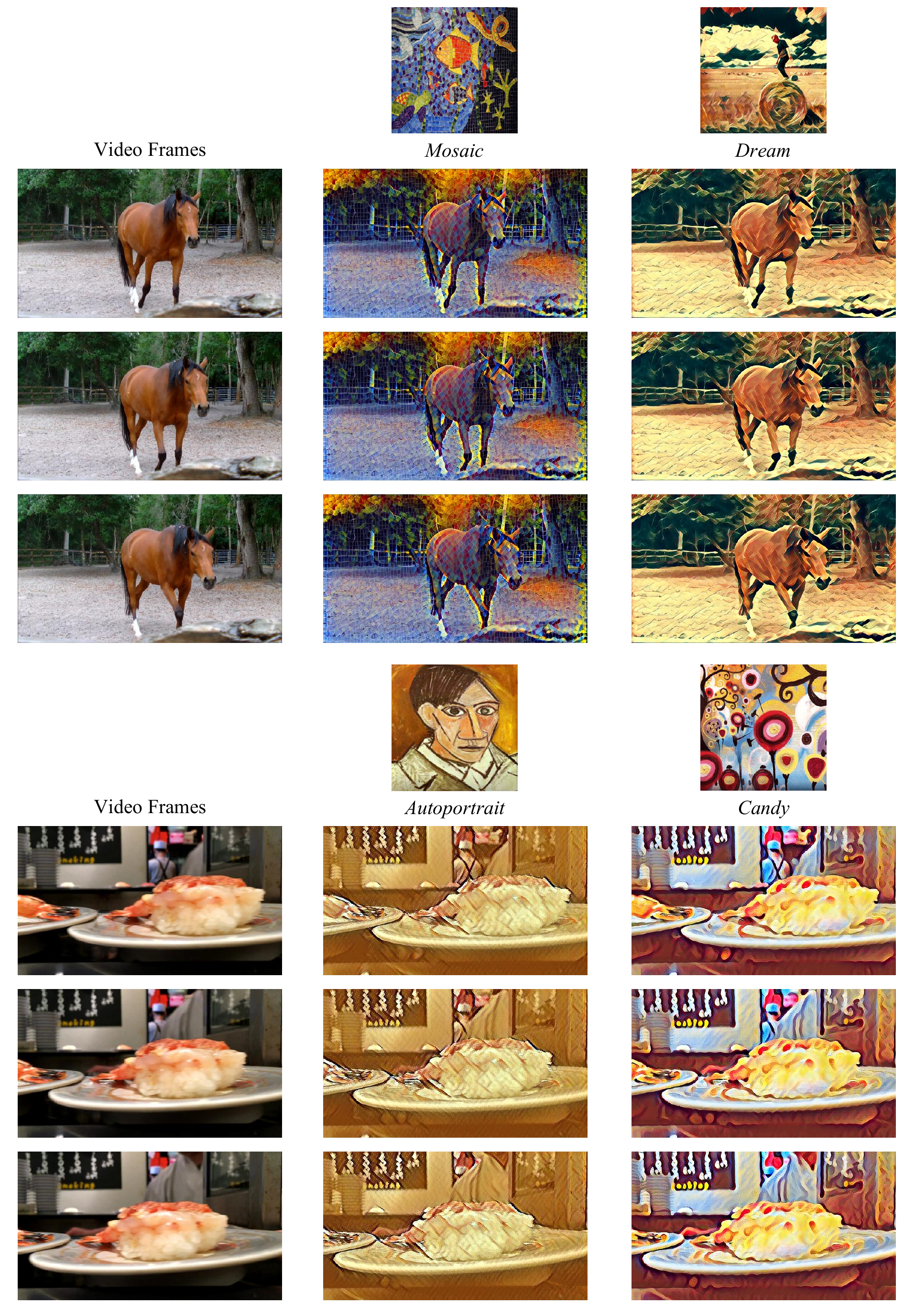}
\caption{Video style transfer results using ReCoNet. The first column contains two groups of three consecutive image frames in videos downloaded from Videvo.net~\cite{videvonet}. Each video frames are followed by two style target images and their corresponding stylized results of the video frames. The styles are \textit{Mosaic}, \textit{Dream}, \textit{Autoportrait} (Picasso, 1907), and \textit{Candy}}
\label{styles}
\end{figure}

We use Monkaa and FlyingThings3D in the Scene Flow datasets~\cite{mayer2016large} as the training dataset, and MPI Sintel dataset~\cite{butler2012naturalistic} as the testing dataset. The Scene Flow datasets provide optical flows and motion boundaries for each consecutive frames, from which we can also obtain occlusion masks using the method provided by Sundaram~\etal~\cite{sundaram2010dense}. Monkaa dataset is extracted from the animation movie Monkaa and contains around 8640 frames, resembling MPI Sintel dataset. FlyingThings3D dataset is a large dataset of everyday objects flying along random 3D trajectories and contains around 20150 frames, resembling animated and real-world complex scenes. Same as the verification process of previous works~\cite{chen2017coherent,gupta2017characterizing,huang2017real}, we use MPI Sintel dataset to verify the temporal consistency and perceptual styles of our stylization results.

All image frames are resized to $640 \times 360$. We train the model with a batch size of 2 for 30,000 steps, roughly two epochs over the training dataset. We pair up consecutive frames for the two-frame synergic training and adopt random horizontal flip on each pair. The frame pairs are shuffled in training process. We use Adam optimizer~\cite{kingma2015adam} with a learning rate of $ 10^{-3}$, and set the default training hyper-parameters to be $\alpha = 1, \beta = 10, \gamma = 10^{-3}, \lambda_f = 10^7, \lambda_o = 2 \times 10^3$. We implement our style transfer pipeline on PyTorch 0.3~\cite{paszkepytorch} with cuDNN 7~\cite{chetlur2014cudnn}. All tensor calculations are performed on a single GTX 1080 Ti GPU. Further details of the training process can be found in our supplementary materials.

We also download 50 videos from Videvo.net~\cite{videvonet} to verify our generalization capacity on videos in real world. Figure \ref{styles} shows style transfer results of four different styles on three consecutive video frames. We observe that the color, strokes and textures of the style target can be successfully reproduced by our model, and the stylized frames are visually coherent.

\subsection{Comparison to Methods in the Literature} \label{sec:comparison}

\subsubsection{Quantitative Analysis}

\setlength{\tabcolsep}{4pt}
\begin{table}[t]
\caption{Temporal error $e_{stab}$ and average FPS in the inference stage with style \textit{Candy} on different models. Five scenes from MPI Sintel Dataset are selected for validation}
\begin{center}
\begin{tabular}{| c || c | c | c | c | c || c |}\hline
Model & Alley-2 & Ambush-5 & Bandage-2 & Market-6 & Temple-2 & FPS\\\hline
Chen~\etal~\cite{chen2017coherent} & 0.0934 & 0.1352 & 0.0715 & 0.1030 & 0.1094 & 22.5\\\hline 
ReCoNet & 0.0846 & 0.0819 & 0.0662 & 0.0862 & 0.0831 & 235.3\\\hline
Huang~\etal~\cite{huang2017real} & 0.0439 & 0.0675 & 0.0304 & 0.0553 & 0.0513 & 216.8 \\\hline
Ruder~\etal~\cite{ruder2016artistic} & 0.0252 & 0.0512 & 0.0195 & 0.0407 & 0.0361 & 0.8\\\hline
\end{tabular}
\end{center}
\label{quant_experiment}
\end{table}
\setlength{\tabcolsep}{1.4pt}

Table \ref{quant_experiment} shows the temporal error $e_{stab}$ of four video style transfer models on five scenes in MPI Sintel Dataset with style \textit{Candy}. $e_{stab}$ is the square root of output-level temporal error over one whole scene:
\begin{gather}
e_{stab} = \sqrt{ \frac{1}{T-1} \sum_{t=1}^{T} \frac{1}{D} M_t \lVert O_t - W_t(O_{t-1}) \rVert^2 }
\end{gather}
where $T$ is the total number of frames. Other variables are identical to those in the output-level temporal loss. This error function verifies the temporal consistency of traceable pixels in the stylized output. All scene frames are resized to $640 \times 360$. We use a single GTX 1080 Ti GPU for computation acceleration.

From the table, we observe that Ruder~\etal~\cite{ruder2016artistic}'s model is not suitable for real-time usage due to low inference speed, despite its lowest temporal error among all models in our comparison. Among the rest models which reach the real-time standard, our model achieves lower temporal error than Chen~\etal~\cite{chen2017coherent}'s model, primarily because of the introduction of the multi-level temporal loss. Although our temporal error is higher than Huang~\etal~\cite{huang2017real}'s model, our model is capable of capturing strokes and minor textures in the style image while Huang~\etal's model could not. Please refer to the qualitative analysis below for details. 

Another finding is that ReCoNet and Huang~\etal's model achieve far better inference speed than the others. Compared with recurrent models~\cite{chen2017coherent,gupta2017characterizing,ruder2018artistic}, feed-forward models are easier to be accelerated with parallelism since the current iteration do not need to wait for the previous frame to be fully processed.

\subsubsection{Qualitative Analysis}
We examine our style transfer results qualitatively with other real-time models proposed by Chen~\etal~'s~\cite{chen2017coherent} and Huang~\etal~'s~\cite{huang2017real}.

\begin{figure}[t]
\centering
\includegraphics[width=11cm]{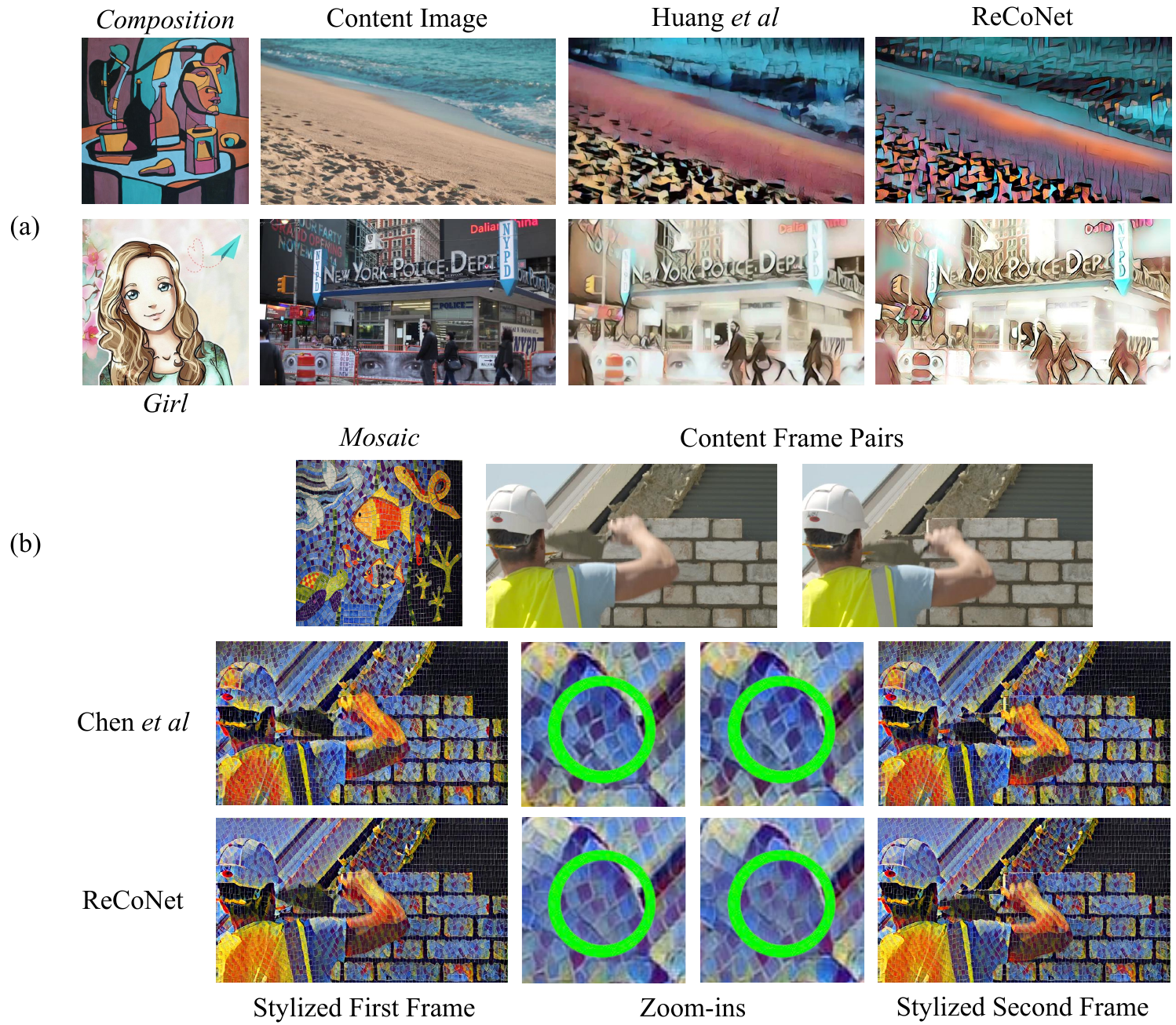}
\caption{Qualitative comparison of style transfer results in the literature. (a) Style transfer results between Huang~\etal~\cite{huang2017real}'s model and ReCoNet on image frames. (b) Style transfer results between Chen~\etal~\cite{chen2017coherent}'s model and ReCoNet on consecutive image frames with zoom-ins of flickering regions}
\label{vsliterature}
\end{figure}

\setlength{\tabcolsep}{4pt}
\begin{table}[t]
\caption{User study result. In each of the two comparisons, we aggregate the results of all four video clips for the three questions. ``Same'' means the voter find that results of both models are similar to each other or it is hard to support one against another}
\begin{center}
\begin{tabular}{| c | c | c | c || c | c | c | c |}\hline
Models & Q1 & Q2 & Q3 & Models & Q1 & Q2 & Q3\\\hline
ReCoNet & 64 & 162 & 152 & ReCoNet & 164 & 42 & 115\\\hline
Chen~\etal~\cite{chen2017coherent} & 64 & 15 & 23 & Huang~\etal~\cite{huang2017real} & 22 & 91 & 42\\\hline
Same & 72 & 23 & 25 & Same & 14 & 67 & 43 \\\hline
\end{tabular}
\end{center}
\label{user_study_result}
\end{table}
\setlength{\tabcolsep}{1.4pt}

Figure \ref{vsliterature}(a) shows the stylization comparison between Huang~\etal~'s model and ReCoNet. Although Huang~\etal~'s model achieves low temporal error quantitatively and is able to capture the color information in the style image, it fails to learn much about the perceptual strokes and patterns. There are two reasons that may account for their weak perceptual styles as shown in the two examples in the figure. First, they use a low weight ratio between perceptual losses and temporal loss to maintain temporal coherence, which brings obvious reduction to the quality of output style. However, in ReCoNet, the introduction of the new temporal losses makes it possible to maintain temporal coherence with a larger perceptual to temporal losses ratio, leading to better preserved perceptual styles. As shown in the first example, our stylized image reproduces the distinct color blocks in the \textit{Composition} style much better than Huang~\etal's result, especially on the uneven sand surfaces and the sea wave. Second, Huang~\etal's model uses feature maps from a deeper layer $relu4\_2$ in the loss network to calculate the content loss, which is difficult to capture low-level features such as edges. In the second example, although sharp bold contours are characteristic in the \textit{Girl} image, their model fails to clearly reproduce such style. Unlike Huang~\etal~'s model, as shown in Figure \ref{coherence} and \ref{vsliterature}(b), Chen~\etal~'s work can well maintain the perceptual information of both the content image and the style image. However, from zoom-in regions, we can find noticeable inconsistency in their stylized results, which can also be quantitatively validated by its high temporal errors.

To further compare our video style transfer results with these two models, we conducted a user study. For each of the two comparisons (ReCoNet vs Huang~\etal~'s and ReCoNet vs Chen~\etal~'s), we chose 4 different styles on 4 different video clips downloaded from Videvo.net~\cite{videvonet}. We invited 50 people to answer (Q1) which model perceptually resembles the style image more, regarding the color, strokes, textures, and other visual patterns; (Q2) which model is more temporally consistent such as fewer flickering artifacts and consistent color and style of the same object; and (Q3) which model is preferable overall. The voting results are shown in Table \ref{user_study_result}. Compared with Chen~\etal~'s model, our model achieves much better temporal consistency while maintaining good perceptual styles. Compared with Huang~\etal~'s model, our results are much better in perceptual styles and the overall feeling although our temporal consistency is slightly worse. This validates our previous qualitative analysis. Detailed procedures and results of the user study can be found in our supplementary materials.


\subsection{Ablation Study} \label{sec:ablation}

\subsubsection{Temporal Loss on Different Levels}

\setlength{\tabcolsep}{4pt}
\begin{table}[t]
\caption{Temporal error $e_{stab}$ with style \textit{Candy} for different temporal loss settings in ReCoNet. Five scenes from MPI Sintel Dataset are selected for validation}
\begin{center}
\begin{tabular}{|c || c | c | c | c | c || c |}\hline
Loss Levels & Alley-2 & Ambush-5 & Bandage-2 & Market-6 & Temple-2 & Average\\\hline
Feature-map only & 0.1028 & 0.1041 & 0.0752 & 0.1062 & 0.0991 & 0.0975\\\hline 
Output only & 0.0854 & 0.0840 & 0.0672 & 0.0868 & 0.0820 & 0.0813\\\hline
Both & 0.0846 & 0.0819 & 0.0662 & 0.0862 & 0.0831 & 0.0804\\\hline
\end{tabular}
\end{center}
\label{temporal_loss_levels}
\end{table}
\setlength{\tabcolsep}{1.4pt}

\begin{figure}[t]
\centering
\includegraphics[width=11cm]{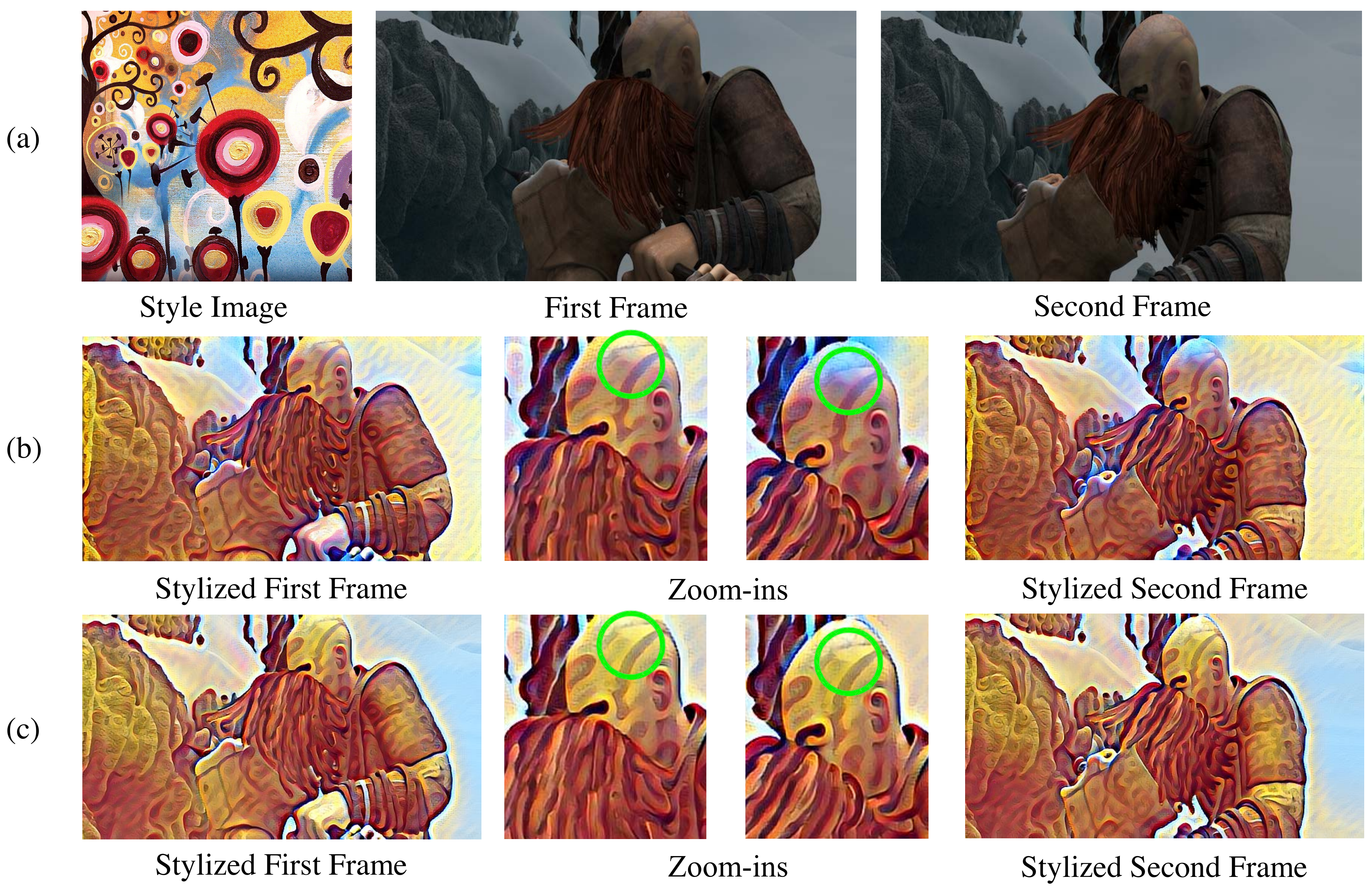}
\caption{Temporal inconsistency in traceable objects. (a) The style target and two consecutive frames in MPI Sintel Dataset. (b) Stylized frames generated without feature-map-level temporal loss. (c) Stylized frames generated with feature-map-level temporal loss. A specific traceable region is circled for comparison}
\label{featout}
\end{figure}

To study whether the multi-level temporal loss does help reduce temporal inconsistency and maintain perceptual style, we implement our video style transfer model on \textit{Candy} style with three different settings: feature-map-level temporal loss only, output-level temporal loss only, and feature-map-level temporal loss plus output-level temporal loss.

Table \ref{temporal_loss_levels} shows the temporal error $e_{stab}$ of these settings on five scenes in MPI Sintel Dataset. We observe that the temporal error is greatly reduced with the output-level temporal loss, while the feature-map-level temporal loss also improves temporal consistency on average.

Figure \ref{featout} demonstrates a visual example of object appearance inconsistency. When only using output-level temporal loss, the exactly same object may alter its color due to the changes of surrounding environment. With the feature-map-level temporal loss, features are preserved for the same object.

\subsubsection{Luminance Difference}

\begin{figure}[t]
\centering
\includegraphics[width=12cm]{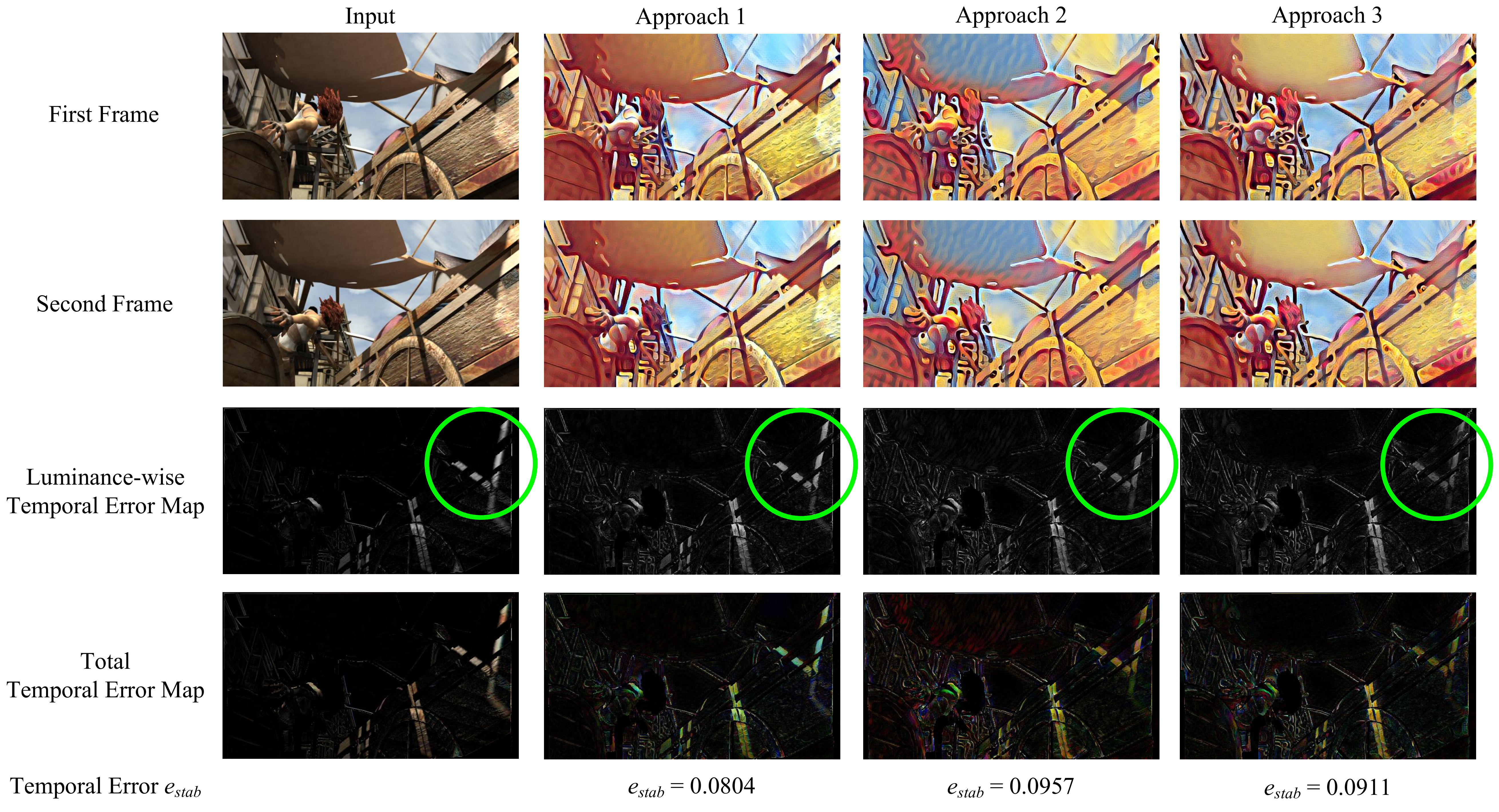}
\caption{Style transfer results using three different approaches described in Section \ref{sec:ablation} to target luminance difference. The style target is \textit{Candy}, and the validation scenes are same as Table \ref{temporal_loss_levels} for temporal error $e_{stab}$ calculation. The total and the luminance-wise temporal error maps show the absolute value of temporal errors in all color channels and in the relative luminance channel respectively}
\label{luminance}
\end{figure}

We compare three different approaches taking or not taking luminance difference into consideration at the output level: 

\begin{enumerate}
\item A relative luminance warping constraint on each RGB channel (Formula \ref{eq:temp_o});
\item Change color space of output-level temporal loss into XYZ color space, then add a relative luminance warping constraint to Y channel: $\mathcal{L}_{temp}^o = \frac{1}{D} M_t ( \lVert (O_t - W_t(O_{t-1}))_{Y} - (I_t - W_t(I_{t-1}))_{Y} \rVert_2+ \lVert (O_t - W_t(O_{t-1}))_{X, Z}\rVert_2)$ where $X, Y, Z$ are the XYZ channels;
\item No luminance constraint: $\mathcal{L}_{temp}^o = \frac{1}{D} M_t \lVert (O_t - W_t(O_{t-1}))_{R,G,B}\rVert_2$.
\end{enumerate}

Other variables in approach 2 and 3 are identical to those in Formula \ref{eq:temp_o}. As shown in Figure \ref{luminance}, all three approaches can obtain pleasant perceptual styles of \textit{Candy} despite some variations in color. However, the first approach has a more similar luminance-wise temporal error map to the input frames compared with the other two methods, especially in the circled illuminated region. This shows the first approach can preserve proper luminance changes between consecutive frames as those in the input, and therefore leads to more natural stylizing outputs. Moreover, the total temporal error map of the first approach is also closer to zero than the results of other two approaches, implying more stable stylized results. This is also supported numerically by a much lower overall temporal error produced by the first approach in the validation scenes. Based on both qualitative and quantitative analysis, we can conclude that adding a relative luminance warping constraint to all RGB channels can generate smoother color change on areas with illumination effects and achieve better temporal coherence.

\section{Conclusions}
In this paper, we present a feed-forward convolutional neural network \textit{ReCoNet} for video style transfer. Our model is able to generate coherent stylized videos in real-time processing speed while maintaining artistic styles perceptually similar to the style target. We propose a luminance warping constraint in the output-level temporal loss for better stylization stability under illumination effects. We also introduce a feature-map level temporal loss to further mitigate temporal inconsistency. In future work, we plan to further investigate the possibility of utilizing both chromaticity and luminance difference in inter-frame warping results for better video style transfer algorithm.

\clearpage
\bibliographystyle{splncs}
\bibliography{egbib}

\end{document}